\pdfoutput=1

\documentclass[final]{cvpr}

\usepackage{times}
\usepackage{epsfig}
\usepackage{graphicx}
\usepackage{amsmath}
\usepackage{amssymb}
\usepackage{booktabs}
\usepackage{xcolor}
\usepackage{microtype}
\usepackage{caption} 
\usepackage[pagebackref=true,breaklinks=true,colorlinks,bookmarks=false]{hyperref}

\begin{document}

%%%%%%%%% TITLE
\title{Toward Transformer-Based Object Detection}

\author{Josh Beal\thanks{Authors contributed equally.}\qquad Eric Kim\footnotemark[1]\qquad Eric Tzeng \qquad Dong Huk Park \qquad Andrew Zhai \qquad Dmitry Kislyuk\\
Pinterest\\
{\tt\small \{jbeal, erickim, etzeng, dhukpark, andrew, dkislyuk\}@pinterest.com}
}

\maketitle

%%%%%%%%% ABSTRACT
\begin{abstract}
Transformers have become the dominant model in natural language processing, owing to their ability to pretrain on massive amounts of data, then transfer to smaller, more specific tasks via fine-tuning.
The Vision Transformer was the first major attempt to apply a pure transformer model directly to images as input, demonstrating that as compared to convolutional networks, transformer-based architectures can achieve competitive results on benchmark classification tasks.
However, the computational complexity of the attention operator means that we are limited to low-resolution inputs.
For more complex tasks such as detection or segmentation, maintaining a high input resolution is crucial to ensure that models can properly identify and reflect fine details in their output.
This naturally raises the question of whether or not transformer-based architectures such as the Vision Transformer are capable of performing tasks other than classification.
In this paper, we determine that Vision Transformers can be used as a backbone by a common detection task head to produce competitive COCO results.
The model that we propose, ViT-FRCNN, demonstrates several known properties associated with transformers, including large pretraining capacity and fast fine-tuning performance. We also investigate improvements over a standard detection backbone, including superior performance on out-of-domain images, better performance on large objects, and a lessened reliance on non-maximum suppression.
We view ViT-FRCNN as an important stepping stone toward a pure-transformer solution of complex vision tasks such as object detection.
\end{abstract}

%%%%%%%%% BODY TEXT
\section{Introduction}

The Transformer~\cite{vaswani2017attention} model has become the preferred solution for a wide range of natural language processing (NLP) tasks, showing impressive progress in machine translation~\cite{lewis2019bart}, question answering~\cite{devlin2018bert}, text classification~\cite{raffel2019exploring}, document summarization~\cite{zhang2019hibert}, and more. Part of this success comes from the Transformer's ability to learn complex dependencies between input sequences via self-attention, and its scalability that makes it possible to pretrain models of remarkable size on large datasets with no signs of saturating performance~\cite{devlin2018bert,radford2018improving,radford2019language,brown2020language}.

The Vision Transformer (ViT)~\cite{dosovitskiy_arxiv2020} demonstrated for the first time that a transformer architecture can be directly applied to images as well, by treating an image as a sequence of patches.
Although its performance on mid-sized datasets trails behind convolution-based models, the ViT seems to retain the capacity seen in NLP transformers, enabling it to pretrain on an unprecedented amount of data.
In effect, ViT suggests that the standard convolution, which has been the hallmark of vision modeling for decades, may be supplemented or replaced by attention-based components.

For convolutional networks, the locality of the convolution operation means that feature maps at higher layers of the network spatially reflect the input---for example, features generated by objects on the left of an input image tend to appear on the left of higher level feature maps as well.
This property makes it relatively straightforward to generalize convolutional classifiers into object detectors by simply feeding high level convolutional maps into detection heads that output object classes and locations.
In contrast, transformers are capable of globally attending at every layer of the network, potentially making the spatial correspondence between the input and intermediate features weaker.
This naturally raises the question of whether or not Vision Transformers can be fine-tuned to perform tasks that are more locally-sensitive, such as object detection or segmentation.

We propose a new model, ViT-FRCNN, that attempts to answer this question by augmenting a ViT with detection-specific task heads to detect and localize objects in images. Importantly, ViT-FRCNN demonstrates that a transformer-based backbone can retain sufficient spatial information for object detection.
We show that ViT-FRCNN achieves competitive results on the COCO detection challenge~\cite{lin2014microsoft}, while exhibiting many of the desirable properties of transformer-based models.
In particular, our experiments suggest that object detection tasks benefit from the massive pretraining paradigm commonly used with transformers. Our experiments also find improved detection performance on large objects (perhaps due to the ability of the architecture to attend globally), and fewer spurious overdetections of objects.

We believe ViT-FRCNN shows that the commonly applied paradigm of large scale pretraining on massive datasets followed by rapid fine-tuning to specific tasks can be scaled up even further in the field of computer vision, owing to the model capacity observed in transformer-based architectures and the flexible features learned in such backbones.
\section{Related work}

\paragraph{Self-attention for object detection:} DETR~\cite{carion_arxiv2020} is notable in that it is the first approach to successfully utilize transformers for the object detection task. 
Specifically, DETR added a transformer encoder and decoder on top of a standard CNN model (e.g., ResNet-50/101), and uses a set-matching loss function.
A notable property in their approach is that it does not need to use non-maximum suppression (NMS) as a post-processing step, as their decoder architecture learns to self-suppress duplicate bounding box predictions.

Deformable DETR~\cite{zhu2020deformable} improves upon some of the limitations exhibited by DETR: 1) requiring much longer training epochs than typical detectors for convergence and 2) achieving relatively low detection performance on small objects. The shortcomings mainly stem from prohibitive complexities in processing high-resolution feature maps and Deformable DETR addresses this by introducing a deformable attention module that learns to attend to small set of sampling locations in the feature map. RelationNet++ \cite{relationnetplusplus2020} leverages multi-head attention in a different manner, proposing a ``Bridging Visual Representations'' (BVR) module based on a Transformer decoder for integrating information from different object representations. The above works build on prior work by RelationNet~\cite{hu2018relation} and Non-Local Networks (NL)~\cite{wang2018non} in using attention between bounding box features and pixel features, respectively, for object detection.

Both DETR and Deformable DETR rely on convolutional networks to encode visual features while the Transformer is utilized for decoding such features into detection outputs. In this paper, we explore a different variant of Transformer-based detector wherein the Transformer is used to encode visual features instead and a traditional region proposal network (RPN)~\cite{ren2015faster} is employed to output detections.

\paragraph{Self-attention for visual representations:} ViT~\cite{dosovitskiy_arxiv2020} is the first pure Transformer-based visual model to perform comparably to state-of-the-art convolutional networks on image recognition tasks. In particular, ViT demonstrates excellent performance when first pretrained on a large scale dataset and then transferred to tasks with fewer datapoints. Despite its effectiveness for image recognition, it is yet to be shown that the approach can be generalized to spatial tasks such as object detection or image segmentation.  

Image GPT (iGPT)~\cite{chen2020generative} is an unsupervised generative pre-training approach for learning strong visual representations. Applying a GPT-2~\cite{radford2019language} based model directly to the image pixels was shown to yield compelling image completions and samples, showing that a fully Transformer-based architecture is possible for some visual tasks, despite the input image resolution being quite limited in their approach.

Other works have explored the relationship between self-attention modules and convolutional layers \cite{cordonnier2019relationship} as well as the limitations of convolutional neural networks~\cite{zhang_icml2019, hermann2020origins}.

\paragraph{Pretraining and self-training:} Several works have explored the effectiveness of pretraining on large-scale image datasets such as JFT-300M~\cite{sun2017revisiting} and IG-940M~\cite{mahajan2018exploring} for visual representation learning. In Big Transfer (BiT)~\cite{kolesnikov2019big}, large-scale classification-based pretraining was found to be beneficial for detection transfer performance. Other works have found that smaller-scale classification-based pretraining, i.e., pretraining on ILSVRC-2012~\cite{deng2009imagenet}, does not necessarily benefit detection performance relative to training from scratch when the detection dataset is sufficiently large~\cite{he2019rethinking, zoph2020rethinking}. In this work, we focus on pretraining paradigms for object detection; semi-supervised learning~\cite{yalniz2019billion} and self-training~\cite{rosenberg2005semi, zoph2020rethinking} paradigms are also beneficial in the context of leveraging large-scale unlabeled data.

\section{Method}

\begin{figure*}
  \centering
  \includegraphics[width=\textwidth]{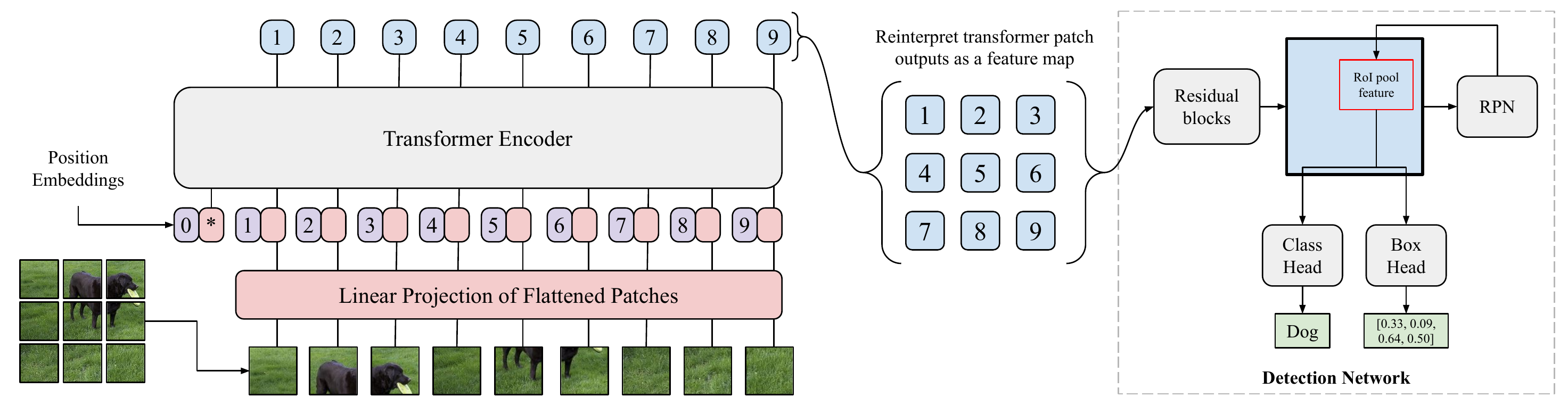}
  \caption{
    We repurpose a Vision Transformer backbone to perform object detection by making use of the per-patch outputs in the final transformer layer.
    By reinterpreting these outputs spatially, we create a feature map that naturally lends itself as the input to a detection model that produces class predictions and regresses to box coordinates.
    The resulting model, ViT-FRCNN, achieves strong performance while exhibiting many of the desirable properties of transformer-based models.
  }
  \label{fig:architecture}
\end{figure*}

We now describe our model, ViT-FRCNN, which augments a Vision Transformer backbone with a detection network so that it can produce bounding box classifications and coordinates.
In doing so, we demonstrate that the ViT is capable of transferring representations learned for classifications to other tasks such as object detection, paving the path for a general class of transformer-based vision models.

Because the ViT is primarily concerned with classification, it uses only the state corresponding to the input class token at the final transformer layer as the final feature to be fed into the classification head.
The remaining tokens in the sequence are used only as features for the final class token to attend to.
However, these unused outputs correspond to the input patches, and in theory could encode local information useful for performing object detection.
We thus propose reinterpreting the final transformer states, excluding the class token, as a spatial feature map.

This feature map is then fed to a detection network modeled after Faster R-CNN~\cite{ren2015faster}.
Like in Faster R-CNN, first a region proposal network (RPN) densely predicts the presence or absence of objects throughout the image.
The features corresponding to the top region proposals are then RoI pooled and fed into a detection head, which classifies each region into object categories and regresses to precise bounding box coordinates.
The ViT-FRCNN architecture is depicted pictorially in Figure~\ref{fig:architecture}.

The RPN identifies regions of interest likely to contain objects by producing multiple predictions per location on the feature map: each prediction corresponds to a different anchor of varying size and aspect ratio, centered at the location of the feature.
Each prediction consists of a binary classification (object vs. no object) and a regression to box coordinates.
The bounding boxes are predicted as offsets from anchor boxes, using the parameterization
\begin{equation*}
  \begin{aligned}[l]
    t_x &= (x - x_a) / w_a \\
    t_y &= (y - y_a) / h_a \\
    t_w &= \log(w / w_a) \\
    t_h &= \log(h / h_a)
  \end{aligned}
  \qquad
  \begin{aligned}[l]
    t_x^* &= (x^* - x_a) / w_a \\
    t_y^* &= (y^* - y_a) / h_a \\
    t_w^* &= \log(w^* / w_a) \\
    t_h^* &= \log(h^* / h_a),
  \end{aligned}
\end{equation*}
where $x$, $y$, $w$, and $h$ denote the the box center, width, and height, and $x$, $x_a$, $x^*$ correspond to the prediction, anchor, and ground truth, respectively.
The box training loss is simply a Huber loss between the predicted offsets $t$ and the ground truth offsets $t^*$.

After the RPN densely predicts regions likely to contain objects, the top candidates are used to RoI-pool regions from the feature map, generating one feature per region proposal.
These pooled features are then fed to a pair of lightweight heads to produce object category (or background) predictions and bounding box regressions, parameterized in the same way as before.
ViT-FRCNN is fully trainable end-to-end, and we jointly train the RPN and detection heads in a single phase.

One final consideration is that, unlike classification, detection relies heavily on fine detail present in small areas of the input image.
Thus, we train at much higher resolution than ViT in an attempt to maintain as much resolution as possible, subject to the limitations of GPU memory.
We also maintain the aspect ratio of the input image, as is standard practice for detection models.
Both of these changes require special handling of the position embedding in the ViT backbone, which expects square images of a fixed size.
ViT-FRCNN simply bilinearly interpolates the position embeddings at runtime to match any given input size and shape.

\subsection{Implementation details}

We chose to use many of the same hyperparameter settings for ViT-FRCNN as in the original Faster R-CNN.
The RPN predicts 15 anchors at each location: we use 5 areas ($32^2$, $64^2$, $128^2$, $256^2$, and $512^2$ pixels) and 3 aspect ratios (1:1, 1:2, and 2:1).
The region proposals produced by the RPN undergo a round of non-maximum suppression (NMS) with overlap threshold 0.7.
During training, the top 2,000 proposals are fed to the detector heads, and for inference we use the top 1,000.
Finally, at inference time we apply a final round of NMS with threshold 0.5 to produce our final detections.

\begin{figure}
\centering
  \includegraphics[width=0.33\linewidth]{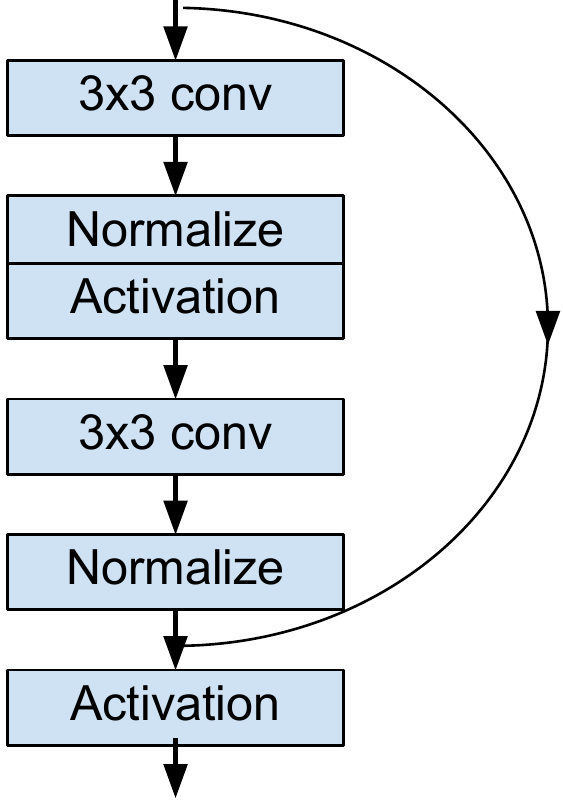}
  \caption{
The intermediate residual convolution blocks \cite{DBLP:journals/corr/HeZRS15} that we use to connect the encoder outputs to the detection module.
In all experiments, we use batch normalization \cite{DBLP:journals/corr/IoffeS15} and GeLU \cite{DBLP:journals/corr/HendrycksG16} activation function.
  }
  \label{fig:resblock}
\end{figure}

We explored two ways of utilizing the encoder outputs: using only the final encoder outputs, or using all intermediate encoder outputs via concatenation.
For both approaches, we reduce the dimensionality of the spatial feature map to 512 channels via a learned transformation.
Finally, in practice we found it helpful to add intermediate residual blocks between the encoder spatial feature map and the detection module.
See Figure~\ref{fig:resblock} for a visualization of the residual block structure, which mimics those of ResNet with minor changes.

\section{Experiments}

We validate the effectiveness of ViT-FRCNN with quantitative results on the benchmark COCO 2017 detection dataset.
We additionally provide a detailed set of analyses and ablations of our model, demonstrating the advantages of a Transformer-based detector for ease of training and out-of-distribution generalization.

Unless otherwise noted, we use the \textit{torchvision}~\cite{10.1145/1873951.1874254} ResNet and Faster R-CNN (FRCNN) implementations in all of our object detection experiments.

\paragraph{Baseline}
As a baseline, we compare against the ResNet-FRCNN-FPN detection model, which is a Faster-RCNN model with Feature Pyramid Networks~\cite{DBLP:journals/corr/LinDGHHB16}.
We trained our own ResNet50-FRCNN-FPN and ResNet101-FRCNN-FPN baselines using \textit{torchvision}'s detection library.

\paragraph{Training hyperparameters}
\begin{table}
\small
    \centering
    \begin{tabular}{lccc}
        \toprule
        Model & LR & Batch Size & Image Size \\
        \midrule
ViT-B/32-FRCNN & 0.032 & 256 & 720/1200 \\ 
ViT-B/32-FRCNN\textsubscript{stride=0.5} & 0.004 & 32 & 720/1200 \\
ViT-B/16-FRCNN & 0.004 & 32 & 672/1092 \\
ViT-B/16*-FRCNN & 0.004 & 32 & 672/1092 \\
        \bottomrule
    \end{tabular}
    \caption{
Hyperparameters for different ViT-FRCNN combinations.
``Image Size'' denotes the target image size and maximum image size respectively. Decreasing the patch size from $32 \times 32$ to $16 \times 16$ considerably improves detection performance, but significantly increases GPU memory usage, resulting in the smaller batch sizes and learning rates.
}
    \label{tab:vit_frcnn_train_hyperparams}
\end{table}

For all ViT-FRCNN experiments, we use SGD with momentum 0.9 and weight decay set to 1e-4.
Due to GPU memory limits, the learning rates, batch sizes, and target/max image size depend on the ViT variant, and are explicitly enumerated Table~\ref{tab:vit_frcnn_train_hyperparams}.
During detection fine-tuning, we do not freeze any layers, and train the entire model end-to-end.

For all ResNet-based experiments, we used SGD with learning rate 0.08, momentum 0.9, weight decay 1e-4, a total batch size of 256, and a target/max image size of 800 and 1333 respectively.
Notably, during detection fine-tuning we freeze the first two ResNet blocks (``stem'', ``res2'').

All experiments use the same train schedule of 21 train epochs, with a learning-rate drop by 10$\times$ at epoch 17.
We always report the detection evaluation metrics using the final model snapshot.
We also used a learning rate linear warm up schedule for the first 1,000 steps, where the initial learning rate is multiplied by 1e-4 and gradually increased to the target learning rate.

During training, we randomly flip the image horizontally with 50\% probability.
Each image is rescaled to a target scale that does not exceed a maximum size.

\paragraph{Pretraining datasets}
In prior work, the Vision Transformer architecture was shown to perform well when pretrained on large-scale image datasets, including ImageNet-21k~\cite{deng2009imagenet} and JFT-300M~\cite{sun2017revisiting}. We investigate the impact of large-scale classification-based pretraining on the detection transfer performance. We pretrained the ViT backbone on ImageNet-21k, a public dataset consisting of 14.2 million images with 21k supervised labels, and Annotations-1.3B, an internal dataset consisting of 1.3 billion images with 18k weakly-supervised labels. The Annotations-1.3B dataset is over 4$\times$ larger in image count than JFT-300M, and achieves competitive ILSVRC-2012 transfer performance when used for large-scale pretraining, as shown in the appendix. For a full comparision of pretraining datasets, we additionally pretrain on ILSVRC-2012, though in Vision Transformers, this dataset was found to be insufficient for pretraining due to the smaller size of the dataset.

\subsection{COCO detection}
We present evaluation results on the COCO 2017 validation set in Table~\ref{tab:coco_detection}.
All ViT-FRCNN variants achieve respectable performance on the detection task, though we observe a number of interesting trends.
The first is that the 16$\times$16 models outperform the 32$\times$32 models by a sizable amount, indicating the importance of preserving fine detail for the detection task.
We provide further analysis of this trend in Section~\ref{sec:ablations}.

Second, we note that across all settings, increasing the size of the pretraining dataset leads to better performance on the downstream object detection task.
For instance, by changing the ViT-B/16 pretraining dataset from ImageNet-21k to Annotations-1.3B, we obtain a +1.2 AP boost (36.6 to 37.8 AP). Likewise, when changing the ViT-B/32 pretraining dataset from ImageNet-21k to Annotations-1.3B, we obtain a +1.6 AP boost (29.3 to 30.9 AP).

These results contrast some results from ViT \cite{dosovitskiy_arxiv2020}, where the lightweight ViT-B/32 model displayed worse ILSVRC-2012 transfer performance when scaling from ImageNet-21k to JFT-300M (81.28\% to 80.73\% top-1 accuracy). We observed a similar trend in the ILSVRC-2012 transfer performance when scaling ViT-B/32 from ImageNet-21k to Annotations-1.3B (81.22\% to 80.82\% top-1 accuracy).

These results suggest that pretraining with images of multiple objects in a scene may harm the classification transfer performance, yet still improve the detection transfer performance. A similar observation was made in a comparison of COCO self-training performance on ILSVRC-2012 and Open Images~\cite{zoph2020rethinking}. Our pretraining results indicate that the ILSVRC-2012 transfer performance is not a fully reliable measure of the representation quality for object detection.

While ViT-FRCNN may not achieve state-of-the-art results on COCO, we believe this signifies an important step forward for transformers in the field of computer vision.
Our results show that transformers trained on classification can successfully be transferred to other related tasks such as object detection, while retaining many of the benefits unique to transformers.
In particular, we are especially excited about ViT-FRCNN's ability to translate representations learned on massive classification datasets into improved performance on downstream detection tasks.
This serves as an important first step toward a class of transformer-based models that, when pretrained on massive quantities of data, can be rapidly fine-tuned for specific vision tasks while requiring relatively little task-specific labeled data.

\begin{table*}
    \centering
    \begin{tabular}{lcccccc}
        \toprule
        Model & AP & AP\textsubscript{50} & AP\textsubscript{75} & AP\textsubscript{S} & AP\textsubscript{M} & AP\textsubscript{L} \\
        \midrule

% These table results are with epoch=20 for ALL models
ResNet50-FRCNN-FPN & 36.0 & 57.7 & 38.4 & 20.8 & 40.0 & 46.2 \\
ResNet101-FRCNN-FPN & 38.8 & 59.9 & 42.0 & 22.2 & 43.0 & 50.9 \\
\midrule
ViT-B/32$\dagger$-FRCNN & 24.8 & 42.3 & 25.0 & 7.3 & 26.3 & 41.1 \\
ViT-B/32-FRCNN & 29.3 & 48.9 & 30.1 & 9.0 & 31.8 & 48.8 \\
ViT-B/32*-FRCNN & 30.9 & 50.5 & 31.7 & 9.7 & 33.7 & \textbf{51.5} \\
\midrule
ViT-B/32-FRCNN\textsubscript{stride=0.5} & 34.5 & 53.4 & 36.8 & 15.6 & 36.9 & \textbf{52.3} \\
\midrule
ViT-B/16-FRCNN & 36.6 & 56.3 & 39.3 & 17.4 & 40.0 & 55.5 \\
ViT-B/16*-FRCNN & 37.8 & 57.4 & 40.1 & 17.8 & 41.4 & \textbf{57.3} \\

% These table results are with best_epoch for each model
% ResNet50-FRCNN-FPN & 36.4 & 57.9 & 38.9 & 20.6 & 40.1 & 47.0 \\
% ResNet101-FRCNN-FPN & 39.0 & 60.1 & 42.0 & 22.4 & 43.3 & 51.2 \\
% \midrule
% ViT-B/32$\dagger$-FRCNN & 25.0 & 42.5 & 25.1 & 7.2 & 26.5 & 41.2 \\
% ViT-B/32-FRCNN & 29.5 & 49.0 & 30.2 & 9.0 & 31.8 & 49.3 \\
% ViT-B/32*-FRCNN & 31.0 & 50.9 & 32.0 & 9.7 & 33.9 & \textbf{51.5} \\
% \midrule
% ViT-B/32-FRCNN\textsubscript{stride=0.5} & 34.6 & 53.6 & 36.8 & 15.8 & 37.3 & \textbf{52.7} \\
% \midrule
% ViT-B/16-FRCNN & 36.9 & 56.4 & 39.7 & 17.6 & 40.3 & 55.8 \\
% ViT-B/16*-FRCNN & 38.0 & 57.7 & 40.4 & 17.6 & 42.0 & \textbf{57.8} \\

        \bottomrule
    \end{tabular}
    \caption{
Average Precision (AP) on the COCO val2017 set. 
``B'' denotes the ``ViT-Base'' backbone. 
``*'' denotes that the backbone was pretrained on Annotations 1.3B prior to detection fine-tuning. ``$\dagger$'' denotes that the backbone was pretrained on ILSVRC2012 prior to detection fine-tuning.
All other ViT backbones are pretrained on ImageNet-21k.
Due to GPU memory limits, we are unable to explore a \textit{stride=0.5} variant of the ``ViT-B/16'' architectures.
}
    \label{tab:coco_detection}
\end{table*}

\subsection{Ablations}
\label{sec:ablations}

Because the properties of transformers applied to vision tasks are still relatively poorly understood, we conduct a comprehensive suite of additional analyses and ablations, covering a variety of parameters such as the input patch size, the handling of intermediate encoder features, and the transformer's observed ability to reduce spurious overdetections.

\paragraph{Encoder spatial resolution.}
We found that the encoder's spatial resolution played a critical role in object detection performance.
In Table~\ref{tab:coco_detection}, we consistently see that there is a significant AP gain by moving from a ViT backbone with 32$\times$32 pixel patches to a backbone with 16$\times$16 pixel patches, which doubles the spatial resolution of the encoder.
For instance, for the ViT backbone trained on Annotations-1.3B, we see an absolute AP gain of +6.9 AP by decreasing the patch size from 32$\times$32 to 16$\times$16, achieving 37.8 AP overall. 
This AP gain is accentuated for small objects: 9.7 to 17.8 AP. Increasing the resolution of spatial features is a well-known method to achieve better object detection performance, (i.e. Feature Pyramid Networks~\cite{DBLP:journals/corr/LinDGHHB16}), and we observe that this
carries over into this model architecture as well.

We also investigated whether increasing the spatial resolution of a trained ViT backbone in a post-hoc manner by using overlapping patches can help with the object detection task.
In Table~\ref{tab:coco_detection}, the ViT-B/32-FRCNN\textsubscript{stride=0.5} row is the same backbone as ViT-B/32-FRCNN, but using patches with 50\% overlap.
This is implemented by converting the encoder's first linear projection layer to an equivalent convolution, and setting its stride to half of the patch size.
We see that there is a significant AP boost, particularly for small objects: 29.3 to 34.5 overall, and 9.0 to 15.6 for small objects, indicating that the input handling is an important factor in ensuring the success of ViT-FRCNN.

\paragraph{Number of connecting residual blocks.}
\begin{table*}
    \centering
    \begin{tabular}{lccccccc}
        \toprule
        Model & Res Blocks & AP & AP\textsubscript{50} & AP\textsubscript{75} & AP\textsubscript{S} & AP\textsubscript{M} & AP\textsubscript{L} \\
        \midrule
% with epoch=20
ViT-B/32-FRCNN & 0 & 24.1 & 45.3 & 23.0 & 6.5 & 24.9 & 40.7 \\
ViT-B/32-FRCNN & 4 & 28.8 & 49.0 & 28.8 & 9.4 & 30.9 & 47.4 \\
ViT-B/32-FRCNN & 8 & \textbf{29.3} & 48.9 & 30.1 & 9.0 & 31.8 & 48.8 \\
ViT-B/32-FRCNN & 16 & \textbf{29.3} & 48.8 & 29.8 & 9.1 & 31.6 & 48.7 \\    

% with epoch=best
% ViT-B/32-FRCNN & 0 & 24.2 & 45.4 & 23.0 & 6.4 & 24.9 & 40.9 \\
% ViT-B/32-FRCNN & 4 & 28.9 & 49.0 & 29.5 & 9.3 & 30.8 & 47.6 \\
% ViT-B/32-FRCNN & 8 & \textbf{29.5} & 49.0 & 30.2 & 9.0 & 31.8 & 49.3 \\
% ViT-B/32-FRCNN & 16 & 28.9 & 49.0 & 29.5 & 9.3 & 30.8 & 47.6 \\
        \bottomrule
    \end{tabular}
    \caption{
We investigate the impact of the number of connecting residual blocks between the encoder backbone and the detection module.
Increasing the number of residual connections leads to a significant AP gain, indicating that a transformer pretrained on classification alone is not enough to solve the detection task.
}
    \label{tab:coco_detection_ablate_num_res_blocks}
\end{table*}
In Table~\ref{tab:coco_detection_ablate_num_res_blocks}, we investigate the impact of the number of residual blocks connecting the encoder outputs to the detection module on detection performance.
We find that introducing the residual blocks leads to a significant AP boost from 24.1 AP to 28.8 AP when using 4 blocks, but there are diminishing returns past that number.
Our interpretation is that the encoder outputs require some learned post-processing before their potential can be fully utilized for the detection task.
Unless otherwise noted, we always use 8 blocks for our experiments.

\paragraph{Intermediate encoder outputs.}
\begin{table*}
    \centering
    \begin{tabular}{lccccccc}
        \toprule
        Model & Encoder Outputs & AP & AP\textsubscript{50} & AP\textsubscript{75} & AP\textsubscript{S} & AP\textsubscript{M} & AP\textsubscript{L} \\
        \midrule
% with epoch=20
ViT-B/32-FRCNN & Final & 28.4 & 48.1 & 28.6 & 8.0 & 30.6 & 47.5 \\
ViT-B/32-FRCNN & All & \textbf{29.3} & \textbf{48.9} & \textbf{30.1} & \textbf{9.0} & \textbf{31.8} & \textbf{48.8} \\
        \midrule
ViT-B/16*-FRCNN & Final & 37.6 & 57.1 & 39.6 & 17.2 & 41.1 & 56.8 \\
ViT-B/16*-FRCNN & All & \textbf{37.8} & \textbf{57.4} & \textbf{40.1} & \textbf{17.8} & \textbf{41.4} & \textbf{57.3} \\

% with epoch=best
% ViT-B/32-FRCNN & Final & 28.4 & 48.1 & 28.5 & 8.0 & 30.7 & 47.8 \\
% ViT-B/32-FRCNN & All & \textbf{29.5} & \textbf{49.0} & \textbf{30.2} & \textbf{9.0} & \textbf{31.8} & \textbf{49.3} \\
%         \midrule
% ViT-B/16*-FRCNN & Final & 37.6 & 57.2 & 39.9 & 16.6 & 41.3 & 57.1 \\
% ViT-B/16*-FRCNN & All & \textbf{38.0} & \textbf{57.7} & \textbf{40.4} & \textbf{17.6} & \textbf{42.0} & \textbf{57.8} \\
        \bottomrule
    \end{tabular}
    \caption{
We investigate different ways of utilizing the intermediate encoder features as input to the detection network.
``Final'' indicates that we use only outputs from the final transformer layer, whereas ``All'' indicates that all intermediate transformer states are concatenated.
We observe a small benefit to utilizing the intermediate features, with the effect being more pronounced for the 32$\times$32 model architectures.
}
    \label{tab:coco_detection_ablate_intermediate_enc_feats}
\end{table*}

In Table~\ref{tab:coco_detection_ablate_intermediate_enc_feats}, we investigate different ways of utilizing the intermediate encoder outputs when constructing the spatial feature map used by the detection module.
There is a slight AP gain of 0.2 when concatenating all intermediate encoder outputs instead of using only the final encoder output.
Unless otherwise noted, we always concatenate all intermediate encoder outputs for our experiments.

\paragraph{Overdetections.}

\begin{figure}
  \begin{center}
  \includegraphics[width=\linewidth]{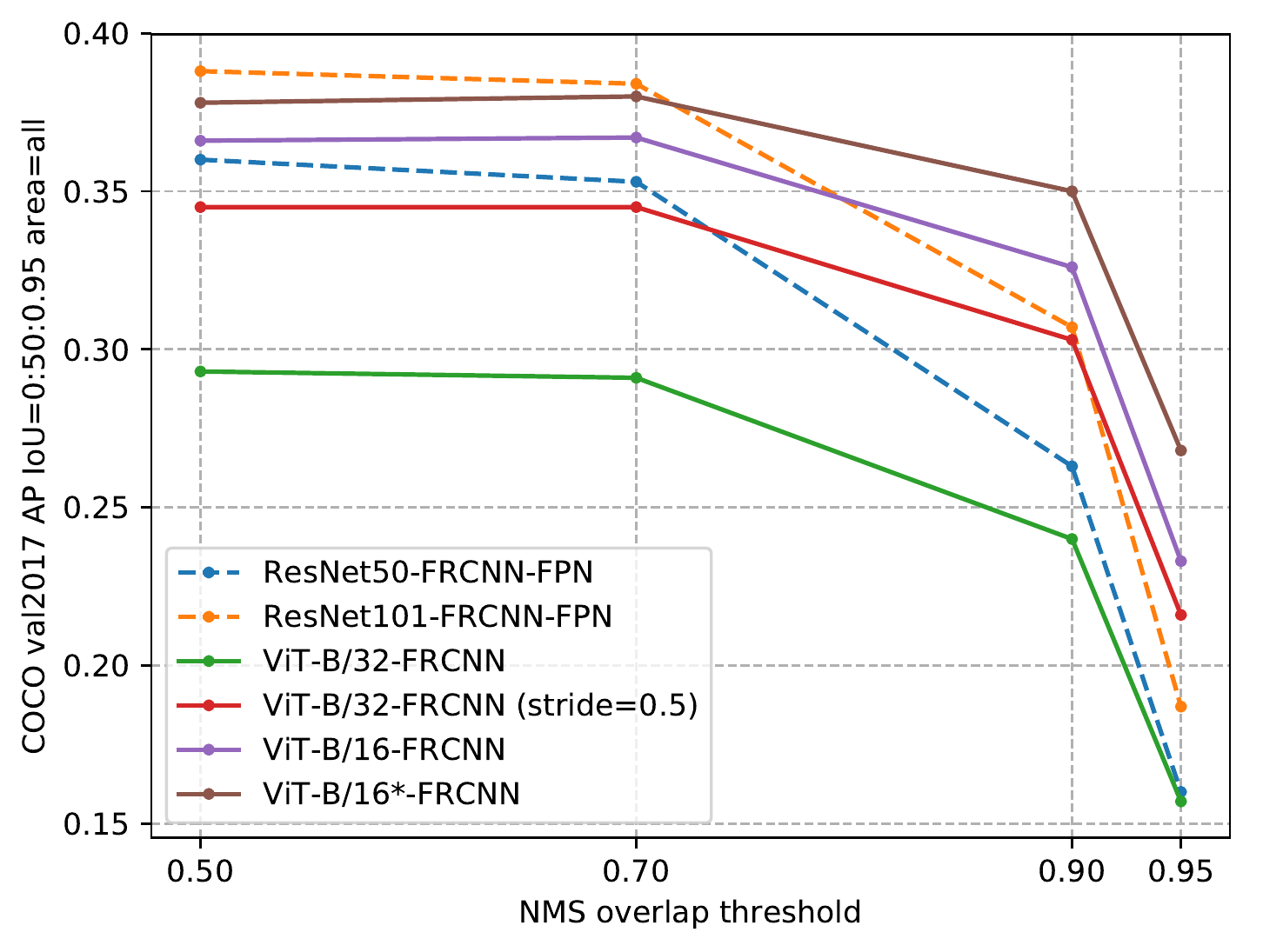}
  \end{center}
  \caption{
    Increasing the non-maximum suppression IoU threshold (thereby decreasing how often suppression occurs) negatively impacts ViT-FRCNN less than the ResNet baselines.
    This indicates that our transformer-based architecture is better able to avoid producing spurious overdetections, possibly due to its ability to have intermediate features attend to each other.
  }
  \label{fig:coco_nms_sensitivity_aps}
\end{figure}

In Figure~\ref{fig:coco_nms_sensitivity_aps}, we investigate whether the ViT-FRCNN architecture is able to avoid producing spurious overdetections of objects.

To investigate this, we evaluated detection models on COCO val2017, and controlled how aggressively the NMS postprocessing will suppress overlapping predicted boxes.
All experiments use the same inference postprocessing parameters: the number of post-NMS detected boxes per image is capped at 100, and predicted boxes with confidence less than 0.05 are discarded.
As the overlap threshold is increased from 0.5 to 0.9, i.e. NMS is nearly disabled, we see that the AP of ResNet-based detection models steeply drops: -9.7 and -8.1 AP for ResNet50 and ResNet101 respectively.
However, the ViT-B/16* detection model has a much smaller drop of -2.8 AP, which suggests that the transformer-based model is able to reduce the occurrences of spurious overdetections.

It is informative to study what happens at the extreme NMS threshold of 0.95, where NMS is effectively disabled.
At the default NMS threshold of 0.5, ResNet101-FRCNN-FPN outperforms ViT-B/16*-FRCNN by 1.0 AP.
However, when the NMS threshold is increased to 0.95, ViT-B/16*-FRCNN significantly outperforms ResNet101-FRCNN-FPN by 8.1 AP.
This quantitatively suggests that the transformer-based detectors produce significantly fewer overdetections than ResNet-based detectors.

\begin{figure}
  \begin{center}
  \includegraphics[width=0.9\linewidth]{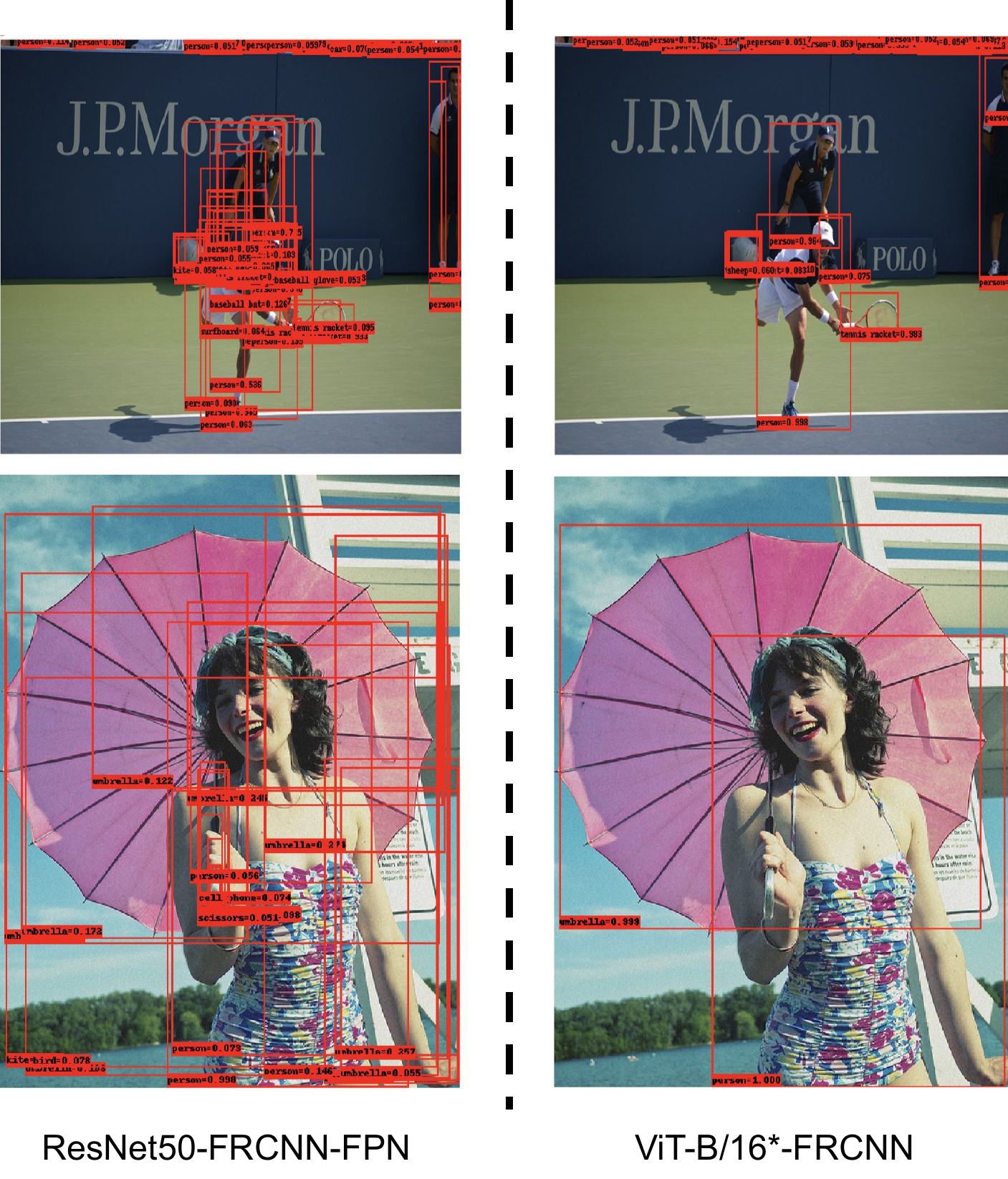}
  \vspace{-0.5cm}
  \end{center}
  \caption{
Comparison of object detection outputs for ResNet-based and transformer-based models.
Left is ResNet50-FRCNN-FPN, right is ViT-B/16*-FRCNN.
The visualized boxes are after NMS is applied with a threshold of 0.5, and boxes with confidence less than 0.05 are discarded.
A trend is that ResNet-FRCNN models generate more spurious boxes for an object (``overdetections'') than ViT-FRCNN models.
  }
  \label{fig:overdetections}
\end{figure}

Qualitatively, in Figure~\ref{fig:overdetections}, we observe that ViT-FRCNN generally avoids predicting multiple spurious boxes that cover different regions of the same object.

The ViT encoder's ability to globally attend to the entire image is a possible explanation for this phenomena.
We speculate that because the transformer's features incorporate both local and global context via the attention mechanism, the FRCNN heads are better able to classify spurious overdetections as the ``background'' category than the ResNet-based baselines.
The underlying mechanism for this behavior warrants further study.

\paragraph{Self-attention maps.}
\begin{table*}
    \centering
    \begin{tabular}{lcccccccc}
        \toprule
        Model & Res Blocks & Concat Attn Maps & AP & AP\textsubscript{50} & AP\textsubscript{75} & AP\textsubscript{S} & AP\textsubscript{M} & AP\textsubscript{L} \\
        \midrule
% with epoch=20
ViT-B/32-FRCNN & 4 & & 28.8 & 49.0 & 28.8 & 9.4 & 30.9 & 47.4 \\
ViT-B/32-FRCNN & 4 & \checkmark & 29.3 & 49.0 & 29.9 & 9.0 & 31.7 & 48.5 \\
        \midrule
ViT-B/32-FRCNN & 8 & & 29.3 & 48.9 & 30.1 & 9.0 & 31.8 & 48.8 \\
ViT-B/32-FRCNN & 8 & \checkmark & 28.8 & 48.8 & 29.3 & 9.0 & 31.1 & 47.3 \\

% with epoch=best
% ViT-B/32-FRCNN & 4 & & 28.9 & 49.0 & 29.5 & 9.3 & 30.8 & 47.6 \\
% ViT-B/32-FRCNN & 4 & \checkmark & 29.6 & 49.0 & 30.1 & 9.3 & 31.8 & 49.0 \\
%         \midrule
% ViT-B/32-FRCNN & 8 & & 29.5 & 49.0 & 30.2 & 9.0 & 31.8 & 49.3 \\
% ViT-B/32-FRCNN & 8 & \checkmark & 29.0 & 48.8 & 29.7 & 9.5 & 31.2 & 47.6 \\
        \bottomrule
    \end{tabular}
    \caption{
%Ablation.
We investigate whether there is any benefit to concatenating the self-attention maps with the encoder outputs prior to the detection modules.
When concatenating attention maps, we first summarize the attention maps by averaging attention maps across all attention heads.
Using attention maps in this manner seems to have a neutral effect on the detection task.
}
    \label{tab:coco_detection_ablate_concat_selfattention}
\end{table*}
We also investigate whether utilizing the encoder self-attention maps as additional features to the detection module helps detection performance.
We accomplish this by concatenating the encoder self-attention maps to the encoder spatial outputs. As there are multiple self-attention heads, we aggregate these attentions into a single attention map via averaging before concatenation.

In Table~\ref{tab:coco_detection_ablate_concat_selfattention}, we see that utilizing the self-attention maps in this way has limited benefit for detection.
We theorize that, because the self-attention is already integrated into the encoder outputs via a dot product, the self-attention maps themselves are redundant for the detection downstream task.

\paragraph{ResNet baselines.}
In our experiments, for simplicity we generally chose to use off-the-shelf detector settings whenever possible, without the addition of any bells and whistles. This also applies to our ResNet-based baseline detectors.
In the appendix, we compare our ResNet-based baselines to other results from the literature, which adopt various tweaks such as additional data augmentation and longer train schedules to improve performance.

\subsection{Curriculum pretraining}
\begin{table*}
    \centering
    \begin{tabular}{lcccccccc}
        \toprule
        Model & IN-21k & OIDV6 & AP & AP\textsubscript{50} & AP\textsubscript{75} & AP\textsubscript{S} & AP\textsubscript{M} & AP\textsubscript{L} \\
        \midrule
% epoch=20
ViT-B/32-FRCNN & \checkmark & & 29.3 & 48.9 & 30.1 & 9.0 & 31.8 & 48.8 \\
ViT-B/32-FRCNN & \checkmark & \checkmark & \textbf{30.4} & \textbf{50.8} & \textbf{31.2} & \textbf{10.2} & \textbf{33.4} & \textbf{49.7} \\
\midrule
ViT-B/32-FRCNN\textsubscript{stride=0.5} & \checkmark & & 34.5 & 53.4 & 36.8 & 15.6 & 36.9 & \textbf{52.3} \\
ViT-B/32-FRCNN\textsubscript{stride=0.5}  & \checkmark & \checkmark & \textbf{34.9} & \textbf{54.6} & \textbf{37.2} & \textbf{16.5} & \textbf{38.1} & \textbf{52.3} \\
% epoch=best
% ViT-B/32-FRCNN & \checkmark & & 29.5 & 49.0 & 30.2 & 9.0 & 31.8 & 49.3 \\
% ViT-B/32-FRCNN & \checkmark & \checkmark & \textbf{30.5} & \textbf{51.1} & \textbf{31.2} & \textbf{10.1} & \textbf{33.8} & \textbf{49.4} \\
% \midrule
% ViT-B/32-FRCNN\textsubscript{stride=0.5} & \checkmark & & 34.6 & 53.6 & 36.8 & 15.8 & 37.3 & \textbf{52.7} \\
% ViT-B/32-FRCNN\textsubscript{stride=0.5}  & \checkmark & \checkmark & \textbf{35.1} & \textbf{55.1} & \textbf{37.5} & \textbf{16.7} & \textbf{38.3} & \textbf{52.7} \\
        \bottomrule
    \end{tabular}
    \caption{
We investigate whether pretraining the transformer backbone on Open Images V6 helps improve detection performance. ``IN-21k'' refers to pretraining on ImageNet-21k, whereas the combination of ``IN-21k'' and ``OIDV6'' refers to pretraining on ImageNet-21k, followed by pretraining on Open Images V6.
All models are then fine-tuned on COCO train2017, and we report results on val2017.
Keeping in line with prior work on transformers, we find that ViT-FRCNN consistently reaps the benefits of larger-scale pretraining.
}
    \label{tab:coco-openimages-pretrain}
\end{table*}
We provide a preliminary investigation of a curriculum pretraining~\cite{bengio2009curriculum, wang2020curriculum} approach for object detection, where the classification-pretrained model is fine-tuned on a larger detection dataset prior to training on COCO 2017.

For this investigation, we consider Open Images V6~\cite{OpenImages}, a supervised dataset consisting of 1.7 million images, with 15.8 million bounding boxes and 600 categories. Relative to COCO 2017 train, which consists of 118k images and 860k bounding boxes, this dataset is an order of magnitude larger in terms of image and bounding box count.
    
To improve the computational efficiency of pretraining, we adopted a simplified detection model architecture inspired by DETR~\cite{carion_arxiv2020}. The CNN-based detector head is replaced by feedforward networks attached to the final encoder layer outputs, which directly output a set of box predictions. 
This model is trained with the set-based global loss and uses fixed box offsets at each encoder output to accelerate convergence in the early phases of training. The appendix contains more details of this model architecture.

The simplified Transformer model is pretrained for 100 epochs on the Open Images V6 dataset, using the AdamW~\cite{loshchilov2018decoupled} optimizer with a base learning rate of 3e-4, weight decay of 0.1, and a total batch size of 4,096. The ViT-B/32 backbone is first pretrained on ImageNet-21k for these curriculum pretraining experiments.

As seen in Table~\ref{tab:coco-openimages-pretrain}, the addition of the pretraining phase on Open Images V6 yields a +1.1 AP improvement for the ViT-B/32-FRCNN model, and a +0.4 AP improvement for the ViT-B/32-FRCNN model with overlapping patches. This phase of pretraining is shown to be most beneficial for improving the performance on small and medium objects.

\subsection{Real-world generalization}

\begin{figure}
  \begin{center}
  \includegraphics[width=\linewidth]{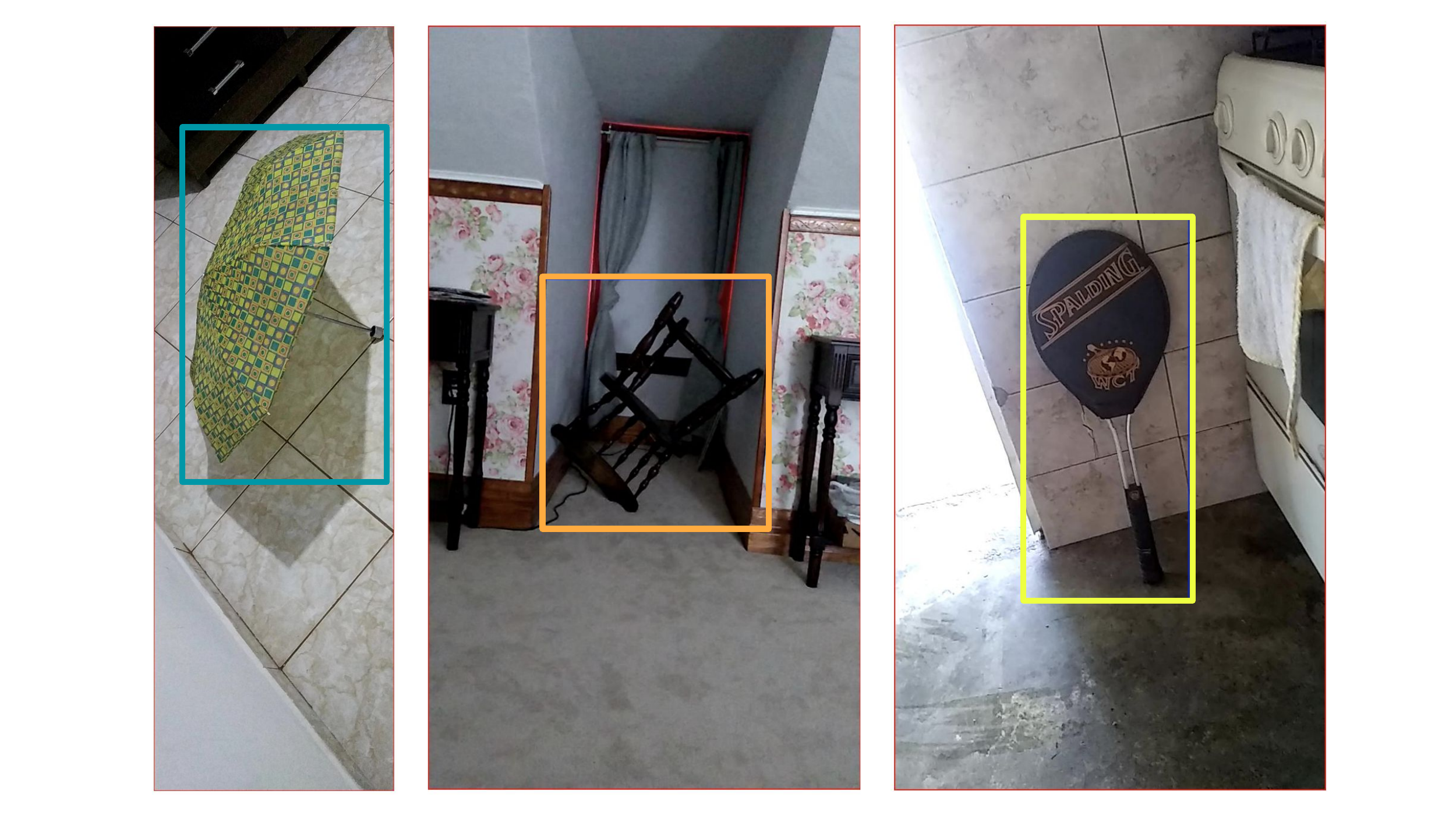}
  \end{center}
  \vspace{-0.5cm}
  \caption{
    Examples from the \textsc{ObjectNet-D} dataset on categories such as umbrellas and tennis rackets, highlighting the controls for rotation, background, and viewpoint.
  }
  \label{fig:objectnet-d}
\end{figure}

ObjectNet~\cite{barbu2019objectnet} is a large-scale test set that measures the generalization performance of vision models in real-world settings by controlling for the biases of rotation, background, and viewpoint in the development of the dataset. ObjectNet has proven to be useful for the evaluation of the generalization and robustness of image classification models~\cite{kolesnikov2019big,djolonga2020robustness,taori2020measuring}. A reanalysis of this dataset provided bounding box annotations for a subset of the original dataset~\cite{borji2020objectnet}.

We construct a novel subset of ObjectNet, referred to as \textsc{ObjectNet-D}, by selecting the COCO 2017 overlapping categories with bounding box annotations that were created in the reanalysis of the dataset. The \textsc{ObjectNet-D} dataset consists of 4,971 test images and 23 object categories. Nearly all of the objects (99.9\%) in this dataset are ``large'' per the object size definitions in COCO 2017. Therefore, we do not focus on the ``small'' and ``medium'' object detection performance in our analysis.  
Figure~\ref{fig:objectnet-d} visualizes a few indicative samples from the \textsc{ObjectNet-D} dataset. This dataset enables a robust test of the detection models with respect to domain shift. The models are evaluated on the corresponding categories without any kind of fine-tuning.

\begin{table}
    \centering
    \begin{tabular}{lcccccc}
        \toprule
        Model & AP & AP\textsubscript{50} & AP\textsubscript{75} & AP\textsubscript{L} \\
        \midrule
% epoch=20
ResNet50-FRCNN-FPN & 13.9 & 30.8 & 10.2 & 14.1 \\
ResNet101-FRCNN-FPN & 15.9 & 34.1 & 12.9 & 16.1 \\
\midrule
ViT-B/32-FRCNN & 16.4 & 34.5 & 14.0 & 16.5 \\
ViT-B/32-FRCNN\textsubscript{stride=0.5} & 17.9 & 36.4 & 15.8 & 18.0 \\
ViT-B/16-FRCNN & 20.2 & 40.5 & 17.8 & 20.4 \\
ViT-B/16*-FRCNN & \textbf{22.9} & \textbf{44.6} & \textbf{20.9} & \textbf{23.0} \\
% epoch=best
% ResNet50-FRCNN-FPN & 14.5 & 31.7 & 11.1 & 14.7 \\
% ResNet101-FRCNN-FPN & 16.9 & 35.6 & 13.9 & 17.1 \\
% \midrule
% ViT-B/32-FRCNN & 17.3 & 35.1 & 15.3 & 17.3 \\
% ViT-B/32-FRCNN\textsubscript{stride=0.5} & 18.5 & 37.2 & 16.1 & 18.6 \\
% ViT-B/16-FRCNN & 20.3 & 40.3 & 18.1 & 20.4 \\
% ViT-B/16*-FRCNN & \textbf{23.1} & \textbf{44.8} & \textbf{21.1} & \textbf{23.2} \\
        \bottomrule
    \end{tabular}
    \caption{
Average Precision (AP) on the \textsc{ObjectNet-D} dataset. ViT-FRCNN yields better out-of-domain performance than ResNet-FRCNN approaches.
Note that we omit the small and medium object size metrics because there are too few examples, leading to uninformative, noisy metrics.
}
    \label{tab:objectnet_detection}
\end{table}
As seen in Table~\ref{tab:objectnet_detection}, ViT-FRCNN models significantly outperform the ResNet-FRCNN baselines on this dataset. ViT-B/16*-FRCNN achieves 22.9 AP (+7 AP boost) relative to the ResNet101-FRCNN-FPN baseline. Increasing the pretraining dataset size results in clear improvements to the out-of-distribution generalization (+2.7 AP boost).

These results suggest that large-scale Transformer pretraining is a promising avenue to improve the performance of detection models in challenging real-world settings.
\section{Conclusion}

In this work, we introduced ViT-FRCNN, a competitive object detection solution which utilizes a transformer backbone, suggesting that there are sufficiently different architectures from the well-studied CNN backbone plausible to make progress on complex vision tasks.
Transformer-based models have demonstrated an ability to pretrain with massive datasets without saturation, and fine-tune to new tasks quickly, both of which are properties we observe with ViT-FRCNN.
We believe that ViT-FRCNN is but the first of many transformer-based architectures that will tackle the wide array of vision problems in the research community.

\section*{Acknowledgements}
We thank Kofi Boakye, Vahid Kazemi, and Chuck Rosenberg for valuable discussions regarding the paper.

{\small
\bibliographystyle{ieee_fullname}
\bibliography{egbib}
}

\renewcommand{\thesection}{\Alph{section}}
\setcounter{section}{0}

\section{Appendix}

\subsection{ResNet baselines}

\begin{table*}
    \centering
    \begin{tabular}{lcccccc}
        \toprule
        Model & AP & AP\textsubscript{50} & AP\textsubscript{75} & AP\textsubscript{S} & AP\textsubscript{M} & AP\textsubscript{L} \\
        \midrule
% epoch=20
ResNet50-FRCNN-FPN\textsubscript{ours} & 36.0 & 57.7 & 38.4 & 20.8 & 40.0 & 46.2 \\
% epoch=best
%ResNet50-FRCNN-FPN\textsubscript{ours} & 36.4 & 57.9 & 38.9 & 20.6 & 40.1 & 47.0 \\
ResNet50-FRCNN-FPN\textsubscript{D2} & 40.2 & 61.0 & 43.8 & 24.2 & 43.5 & 52.0 \\
ResNet50-FRCNN-FPN\textsubscript{+} & 42.0 & 62.1 & 45.5 & 26.6 & 45.4 & 53.4 \\
\midrule
% epoch=20
ResNet101-FRCNN-FPN\textsubscript{ours} & 38.8 & 59.9 & 42.0 & 22.2 & 43.0 & 50.9 \\
% epoch=best
%ResNet101-FRCNN-FPN\textsubscript{ours} & 39.0 & 60.1 & 42.0 & 22.4 & 43.3 & 51.2 \\
ResNet101-FRCNN-FPN\textsubscript{D2} & 42.0 & 62.5 & 45.9 & 25.2 & 45.6 & 54.6 \\
ResNet101-FRCNN-FPN\textsubscript{+} & 44.0 & 63.9 & 47.8 & 27.2 & 48.1 & 56.0 \\
        \bottomrule
    \end{tabular}
    \caption{
Average Precision (AP) on the COCO val2017 set. 
``D2'' denotes the Detectron2 implementation, which has a slightly different ResNet architecture, train-time scale augmentation, and a longer ``3x'' train schedule (37 vs 25 train epochs).
``+'' denotes a heavily-tuned \cite{carion_arxiv2020} implementation: GIoU \cite{DBLP:journals/corr/abs-1902-09630}, random crop train-time augmentation, and a longer ``9x'' training schedule (111 vs 25 train epochs).
}
    \label{tab:coco-resnet-baselines-comparison}
\end{table*}

For completeness, in Table~\ref{tab:coco-resnet-baselines-comparison} we compare our ResNet-based baselines to results from the literature.

We acknowledge that it is possible to substantially improve detection performance of our ResNet-based baselines by adopting the various tweaks in Table~\ref{tab:coco-resnet-baselines-comparison}, such as additional data augmentation and longer train schedules.
However, we feel that because our ViT-FRCNN approaches are trained with the same train settings as our ResNet baselines, it would be inappropriate to compare our ViT-FRCNN results to, say, ResNet101-FRCNN-FPN\textsubscript{+} which utilizes substantially more tweaks to achieve better performance.

\subsection{ImageNet performance}
\begin{table}
    \centering
    \begin{tabular}{lcc}
        \toprule
        Dataset & ViT-B/32 & ViT-B/16 \\
        \midrule
ImageNet & 73.38 & 77.91 \\
ImageNet-21k & 81.28 & 83.97\\
JFT-300M & 80.73 & 84.15\\
Annotations-1.3B & 80.82 & 83.15\\
        \bottomrule
    \end{tabular}
    \caption{
Top-1 Accuracy on ILSVRC-2012 (ImageNet) for different pretraining datasets and ViT model architectures.
}
    \label{tab:imagenet_acc}
\end{table}
Models pretrained on Annotations-1.3B achieve competitive performance when fine-tuned on ILSVRC-2012. Table~\ref{tab:imagenet_acc} contains the full details of ViT-B/32 and ViT-B/16 performance for different pretraining datasets.

\subsection{Curriculum pretraining}
The simplified detection model architecture consists of a Vision Transformer backbone with a box prediction MLP head and class prediction MLP head attached to each encoder output. The configuration of these heads and the set prediction loss function follow the DETR implementation. To improve convergence in the early phases of training, a box prediction offset is provided at each output, where each value is set to the corresponding patch's center location.

\subsection{Object distributions}
It is well-known that the distributions of word frequencies in natural language corpora approximately follow Zipf's law~\cite{zipf1999psycho}. We investigated whether visual objects follow a similar distribution on web scale data. For this analysis, we consider the previously mentioned detection datasets---COCO 2017 and Open Images V6 ---as well as Object Index, a semi-supervised internal dataset consisting of 248 million images, with 646 million bounding boxes and 220 categories.

\begin{figure}[t]
  \begin{center}
  \includegraphics[width=0.9\linewidth]{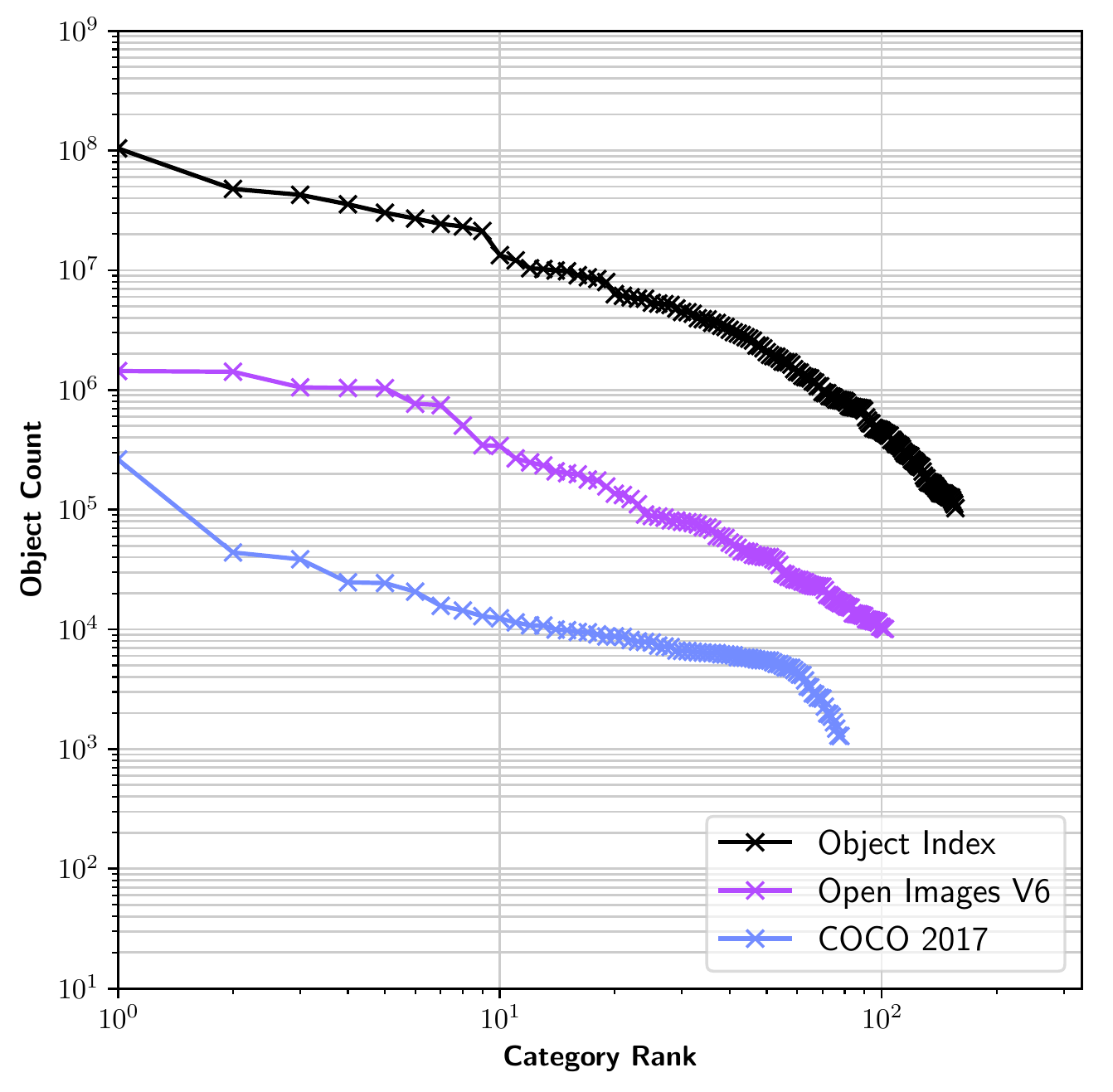}
  \end{center}
  \caption{
    Category distribution of bounding box annotations for object detection datasets of various sizes. The object category distributions converge to a Zipfian distribution as the object detection datasets increase in scale.
  }
  \label{fig:object-distribution}
\end{figure}

Figure~\ref{fig:object-distribution} demonstrates that the distribution of class frequencies for visual objects approximately follows Zipf's law as the detection dataset size increases in scale. This connection to language modeling may be worthy of further investigation as pretraining datasets continue to increase in scale and develop a longer-tailed distribution of categories. In particular, there are well-known procedures for handling class imbalance in training on natural language datasets~\cite{mikolov2013distributed, mahajan2018exploring}.

\subsection{ObjectNet-D categories}

We provide a list of the 29 ObjectNet categories that overlap with ILSVRC-2012 and COCO 2017 categories: alarm clock, backpack, banana, beer bottle, bench, bicycle, cellphone, chair, computer mouse, drinking cup, hair dryer, keyboard, laptop (open), microwave, mixing / salad bowl, mug, orange, pill bottle, pop can, remote control, soup bowl, tennis racket, tie, toaster, TV, umbrella, vase, water bottle, wine bottle.

The \textsc{ObjectNet-D} categories consist only of the 23 corresponding COCO 2017 categories: backpack, banana, bench, bicycle, bottle, bowl, cell phone, chair, clock, cup, hair drier, keyboard, laptop, microwave, mouse, orange, remote, tennis racket, tie, toaster, TV, umbrella, vase.

\end{document}